# Semi-Supervised Fine-Tuning for Deep Learning Models in Remote Sensing Applications

Eftychios Protopapadakis
School of Rural and Surveying Engineering
National Technical University of Athens
Zografou, Greece

Anastasios Doulamis
School of Rural and Surveying Engineering
National Technical University of Athens
Zografou, Greece

Nikolaos Doulamis
School of Rural and Surveying Engineering
National Technical University of Athens
Zografou, Greece

Evangelos Maltezos
School of Rural and Surveying Engineering
National Technical University of Athens
Zografou, Greece

*Abstract*— A combinatory approach of two well-known fields: deep learning and semi supervised learning is presented, to tackle the land cover identification problem. The proposed methodology demonstrates the impact on the performance of deep learning models, when SSL approaches are used as performance functions during training. Obtained results, at pixel level segmentation tasks over orthoimages, suggest that SSL enhanced loss functions can be beneficial in models' performance.

## I. INTRODUCTION

Land cover classification is a widely studied field since the appearance of the first satellite images. In the last two decades, the sensors attached to satellites have evolved in a way that nowadays allows the capture of high-resolution multispectral satellite images. This technological advance made detection/ classification of buildings and other man-made structures from satellite images possible [1]. The automatic identification of buildings in urban areas, using remote sensing data, can be beneficial in many applications including cadastre, urban and rural planning, urban change detection, mapping, geographic information systems, monitoring, housing value and navigation.

Typically, sensory data have the form of RGB, thermal, multispectral or LiDAR images. Available information allows the researchers to mitigate any drawbacks related to occlusions, shadows, and vegetation. It also supports the building detection under different geometric and radiometric diversities, which is a common case in complex scenes. Building detection from 2D images has been achieved using a variety of methods, e.g., through a group of pixels sharing common properties or as an object described by specific features or geometric properties [2].

Deep Convolutional Neural Networks (CNNs) have been considered extremely beneficial for semantic segmentation tasks in multiple remote sensing applications [3]–[6]. Stacked autoencoders or similar deep neural network (DNN) are also used [7], [8], and provide accurate results. A typical Deep Neural Network (DNN) training approach consists of two steps: a) *training per layer* and b) *fine tuning of the entire network*. Training per layer is an unsupervised process exploiting all available data, labeled or not. On the other hand, the fine-tuning approach is limited only to available labelled data instances, that is a supervised process. However, in remote sensing applications, the available training data are only a small portion of the total data entities.

In training per layer each layer learns to reconstruct the input values using fewer computational nodes than the number of input feature values. This is *a compression scheme*; we maintain the input information using fewer neurons. When we stack all these layers, we have a deep architecture, also known as *stacked autoencoders* [9], capable of performing non-linear principal component analysis. The mapped inputs, which lie in a reduced dimension space, can be fed to any classifier. Fine tuning is the process of inducing minor adjustments to the weights of a network. It is used when we have stacked pretrained layers to form a DNN structure, suitable for the problem at hand.

The stacked autoencoders network, along with an additional softmax layer at the end, forms a classifier with exceptionally good performance. The performance, in many cases, bests traditional machine learning approaches [10]. Typically, the entire network is fine-tuned, using backpropagation algorithm, with performance functions either the mean square error (MSE) or cross-entropy. Both performance functions are calculated over the available labeled data instances, i.e., a limited set.

Since the labeled data only consists of a small portion of the total data, someone could understand the impact of using unlabeled data during performance function calculations. Semi supervised learning (SSL) is the machine learning field that can be used for such cases [11]. SSL provides a variety of tools that allow the usage of unlabeled data in conjunction with a small amount of labeled data. These tools are mainly regularizers; i.e. functionals, defined by the user and the SSL assumption(s).

## II. RELATED WORK

The automatic extraction of buildings in urban areas using remote sensing data is an essential task in various applications as mentioned above, such as cadastre, urban and rural planning, mapping, GIS systems, housing value and navigation [12], [13].

Financial support has been provided by the Innovation and Networks Executive Agency (INEA) under the powers delegated by the European Commission through the Horizon 2020 program "PANOPTIS–Development of a decision support system for increasing the resilience of transportation infrastructure based on combined use of terrestrial and airborne sensors and advanced modelling tools", Grant Agreement number 769129.

Building extraction from urban scenes with complex architectural structures still remains challenging due to the inherent artifacts (e.g., shadows, etc.) of the used data (remote sensing) as well as the differences in viewpoint, surrounding environment, complex shape and size of the buildings [5]. This topic is an active research field for more than two decades.

Dependent on the data source employed, building extraction techniques can be classified into three groups: i) the ones that use radiometric information (airborne or satellite imagery data) [14]–[16], ii) the ones that exploit height information (LIDAR) [17], [18] and iii) those that combine both of data sources [19], [20]. The most common problems of using only image information for building detection are: i) the presence of shadows, and ii) the fact that urban objects usually present similar pixel values (e.g., building rooftops vs. roads, or vegetation vs. vegetation on building rooftops).

On the other hand, the use of only 3D data from LIDAR sources, such as LIDAR Digital Surface Models (LIDAR/DSMs), provides estimates of low position accuracy and suffers from local under-sampling, reducing the detection accuracy especially for areas of small buildings [21]. Furthermore, using only DSM, it is difficult to distinguish objects of similar height and morphological characteristics, mainly due to the confusion of the trees (having smooth canopies) with the building rooftops.

To overcome the above limitations, a combination of LIDAR/DSM with image information sources, e.g., combining LIDAR data with orthoimages, is applied to improve the building detection accuracy [5]. However, the limitations of using information from multi-modal sources (e.g., LIDAR and imagery data) are the additional cost of acquisition and processing and the co-registration related issues.

Building detection techniques are bounded, in terms of performance, by the available data. Typically, a vast amount of labeled data is required for training and validation. Towards that direction two approaches gained interest past years: a) SSL and b) tensor-based learning. The former [22] employ graph-based approaches, as in [23] that are not scalable and co-trained or are prone to errors induced by wrong predictions between the models [24]. The latter tensor case is not useful if the amount of data exceeds a threshold [25]. In this paper, we concentrate on SSL learnings to improve classification performance in building detection when the number of available data is limited.

*A. Our contribution*

To tackle the challenges for accuracy and economy in building detection, the present study focuses on the combination of data that are extracted from one sensor, i.e., from camera devices (aerial images combined with the corresponding Dense Image Matching - DIM point clouds [26]) and feeding them to a semi-supervised network scheme. The main contribution lies in the incorporation of all available data, during all steps of the training process, labeled or unlabeled. This is achieved by employing three different SSL approaches: a) graph-based [27], b) performance gain maximization [28] and iii) probabilistic framework [29]. These techniques modify existing loss functions during the fine-tuning phase. Generated DNN outperforms SotA approaches by using less than 20% of the already limited annotated data and many unlabeled instances.

All the techniques SSL approaches operate on the non-linear transformation of the input data, imposed by the autoencoders. As such dimensionality related problems are partially handled [9]. The SSL techniques require no modifications, e.g. custom layers. As such, produced (trained) DNN can be easily utilized by third party applications. SSL limitations, e.g. transductive SSL limitation [30], related to the graph creation, do not apply during network's test phase.

III. PROPOSED METHODOLOGY

In this section, we describe the proposed deep SSL approaches for the detection of buildings from color infrared orthoimages. In our case, we used crowdsourced annotations, which correspond to less than 10% of the actual annotations provided with the utilized datasets.

*A. SSL based performance functions*

Assume a loss function $\mathcal{E}(\cdot)$, e.g., MAE or MSE. Then for a given network topology the loss over the labeled data is set as:

$$\mathcal{E}(\widehat{Y} - Y) = \mathcal{E}\left(\begin{bmatrix}\widehat{y}_1\\ \vdots\\ \widehat{y}_l\end{bmatrix} - \begin{bmatrix}t_1\\ \vdots\\ t_l\end{bmatrix}\right) \quad (1)$$

where $\widehat{y}_i, i = 1, \ldots, l$ is the networks output for input $x_i$, and $t_i$ is the actual target corresponding value. That is, $t_i$ are the annotations (targets) for labeled data. Our approach utilizes and the unlabeled instances by extending Eq. (1) to Eq. (2),

$$\mathcal{E}(\widehat{Y} - Y) = \mathcal{E}\left(\begin{bmatrix}\widehat{y}_1\\ \vdots\\ \widehat{y}_l\\ \widehat{y}_{l+1}\\ \vdots\\ \widehat{y}_n\end{bmatrix} - \begin{bmatrix}t_1\\ \vdots\\ t_l\\ \widehat{t}_{l+1}\\ \vdots\\ \widehat{t}_n\end{bmatrix}\right) \quad (2)$$

where $\widehat{y}_i$ and $\widehat{t}_k$, $k = l + 1, \ldots, n$ are the estimated soft target values, i.e. $\widehat{y}_i, \widehat{t}_k \in \mathbb{R}^c$, where $c$ denotes the number of available classes, over the remaining unlabeled data. In this setup, target value $\widehat{t}_k$ is an estimation for the unlabeled instance $x_k$, provided by any SSL technique, described in next section. At this point we should note that utilization of estimations, as actual targets, is extremely risky; back-propagation adjust the weights so that the outputs match the targets. If targets' values are incorrect the DNN will perform poorly.

*B. Employed SSL techniques*

Three different approaches are considered regarding the exploitation of unlabeled data instances during the fine-tuning step of the DNN: i) graph-based, ii) performance gain maximization and iii) probabilistic.

    *a) Anchor graph*

Anchor graph estimates a labeling prediction function $f: \mathbb{R}^m \to \mathbb{R}$, $m$ denotes the number of feature values, defined on the samples of $X$, by using a subset of $p$ instances: $\mathcal{U} = \{u_k\}_{k=1}^p \subset X_L$, of the labeled data, the label prediction function can be expressed as a convex combination [27]:

$$f(\boldsymbol{x}_i) = \sum_{k=1}^{p} Z_{ik} \cdot g(\boldsymbol{u}_k) \qquad (3)$$

where $Z_{ik}$ denotes sample-adaptive weights, which must satisfy the constraints $\sum_{k=1}^{l} Z_{ik} = 1$ and $Z_{ik} \geq 0$ (convex combination constraints). By defining vectors $\boldsymbol{g}$ and $\boldsymbol{a}$ respectively as $\boldsymbol{g} = [g(\boldsymbol{f}_1), \ldots, g(\boldsymbol{f}_n)]^T$ and $\boldsymbol{a} = [g(\boldsymbol{x}_1), \ldots, g(\boldsymbol{x}_p)]^T$, Eq. (3) can be rewritten as $\boldsymbol{g} = \boldsymbol{Z}\boldsymbol{\alpha}$ where $\boldsymbol{Z} \in \mathbb{R}^{n \times p}$, where $n$ denotes the number of data and $p$ the number of subset $\boldsymbol{\mathcal{U}}$ entries. The designing of matrix $\boldsymbol{Z}$, which measures the underlying relationship between the samples of $\boldsymbol{X}_U$ and samples $\boldsymbol{X}_L$, is based on weights optimization, i.e., non-parametric regression. Thus, the reconstruction for any data point is a convex combination [27] of its closest representative samples.

Nevertheless, the creation of matrix $\boldsymbol{Z}$ is not sufficient for labeling the entire data set, as it does not assure a smooth function $\boldsymbol{g}$. Despite the small labeled set, there is always the possibility of inconsistencies in segmentation; different companies with almost identical attributes are classified differently. the following SSL framework is employed:

$$\min_{A=[\boldsymbol{a}_1,\ldots,\boldsymbol{a}_c]} Q(\boldsymbol{A}) = \frac{1}{2}\|\boldsymbol{Z}\boldsymbol{A} - \boldsymbol{Y}\|_F^2 + \frac{\gamma}{2}\text{trace}(\boldsymbol{A}^T\hat{\boldsymbol{L}}\boldsymbol{A}) \qquad (4)$$

where $\hat{\boldsymbol{L}} = \boldsymbol{Z}^T\boldsymbol{L}\boldsymbol{Z}$ is a memory-wise and computationally tractable alternative of the Laplacian matrix $\boldsymbol{L}$. The matrix $\boldsymbol{A} = [\boldsymbol{a}_1, \ldots, \boldsymbol{a}_c] \in \mathbb{R}^{p \times c}$ is the soft label matrix for the representative samples, in which each column vector accounts for a class. Recall that $c$ denotes the number of classes. The matrix $\boldsymbol{Y} = [\boldsymbol{y}_1, \ldots, \boldsymbol{y}_c] \in \mathbb{R}^{n \times c}$ is a class indicator matrix on ambiguously labeled samples with $Y_{ij} = 1$ if the label $l_i$ of sample $i$ is equal to $j$ and $Y_{ij} = 0$ otherwise.

The Laplacian matrix $\boldsymbol{L}$, is calculated as $\boldsymbol{L} = \boldsymbol{D} - \boldsymbol{W}$, where $\boldsymbol{D} \in \mathbb{R}^{n \times n}$ is a diagonal degree matrix and $\boldsymbol{W}$ is approximated as $\boldsymbol{W} = \boldsymbol{Z}\boldsymbol{\Lambda}^{-1}\boldsymbol{Z}^T$. Matrix $\boldsymbol{\Lambda} \in \mathbb{R}^{p \times p}$ is defined as: $\boldsymbol{\Lambda} = \sum_{i=1}^{n} Z_{ik}$. The solution of the Eq. (4) has the form of:

$$\boldsymbol{A}^* = (\boldsymbol{Z}^T\boldsymbol{Z} + \gamma\hat{\boldsymbol{L}})\boldsymbol{Z}^T\boldsymbol{Y} \qquad (5)$$

Each sample label is, then, given by:

$$\hat{l}_i = \arg\max_{j \in \{1,\ldots,c\}} \frac{\boldsymbol{Z}_i \boldsymbol{a}_j}{\lambda_j} \qquad (6)$$

where $\boldsymbol{Z}_i \in \mathbb{R}^{1 \times p}$ denotes the $i$-th row of $\boldsymbol{Z}$, and factor $\lambda_j = \boldsymbol{1}^T \boldsymbol{Z} \boldsymbol{\alpha}_j$ balances skewed class distributions.

*b) Maximal performance gain*

Assume a set of $b$ semi-supervised regressors (SSRs) applied over the same unlabeled data set $\boldsymbol{X}_U$. The outcome would be $b$ predictions, i.e. $\{\boldsymbol{f}_1, \ldots, \boldsymbol{f}_b\}$, where $\boldsymbol{f}_i = \{f(\boldsymbol{x}_{l+1}), \ldots, f(\boldsymbol{x}_{l+u})\}$, $i = 1, \ldots, b$. Let as, also, denote as $\boldsymbol{f}_0$ the predictions over $\boldsymbol{X}_U$, of a traditional supervised approach. If we know the weight of the individual regressors, $\boldsymbol{\alpha} = [a_o, \ldots, a_b]$, $a_i \geq 0$, the problem formulation has the form:

$$\max_{\boldsymbol{f}} \sum_{i=0}^{b} a_i(\|\boldsymbol{f}_0 - \boldsymbol{f}_i\|^2 - \|\boldsymbol{f} - \boldsymbol{f}_i\|^2) \qquad (7)$$

However, since weights are not known a priori, certain assumptions have to be made. At first, we state that $\boldsymbol{a}$ is from a convex linear set $\mathcal{M} = \{a|\boldsymbol{A}^T\boldsymbol{a} \leq \boldsymbol{b}, \boldsymbol{a} \geq 0\}$, where $\boldsymbol{A}$ and $\boldsymbol{b}$ are task-dependent coefficients. Without further knowledge to determine the weights of individual regressors, one aim to optimize the worst-case performance gain:

$$\max_{\boldsymbol{f}} \min_{\boldsymbol{a} \in \mathcal{M}} \sum_{i=0}^{b} a_i(\|\boldsymbol{f}_0 - \boldsymbol{f}_i\|^2 - \|\boldsymbol{f} - \boldsymbol{f}_i\|^2) \qquad (8)$$

The equation above is concave to $f$ and convex to $a$ and thus it is recognized as saddle-point convex-concave optimization [31]. As described in recent work [28] Eq. (8) can be formulated as a geometric projection problem, handling that way the computational load. Specifically, by setting the derivative of Eq. (8) to zero, we get a close form solution w.r.t. $\boldsymbol{f}$ as $\boldsymbol{f} = \sum_{i=1}^{b} a_i\boldsymbol{f}_i$, which we can substitute in Eq. (8), obtaining an expression that is only related to $\boldsymbol{\alpha}$: $\min_{\boldsymbol{a} \in \mathcal{M}}\|\sum_{i=1}^{b} a_i\boldsymbol{f}_i - \boldsymbol{f}_0\|$. As such, we have we have a convex quadratic problem.

*c) information maximization principle*

The Information Maximization Principle (IMP) is a probabilistic classifier is trained in an unsupervised manner, so that a given information measure between data and cluster assignments is maximized. The proposed squared-loss mutual information regularization (SMIR) model [29] is convex (under mild conditions) and results in globally optimal solution.

Let $\mathcal{X} \subseteq \mathbb{R}^d$ and $\mathcal{Y} = \{1, \ldots, c\}$, where $d, c \in Z^+$. Any pair $(\boldsymbol{x}, y) \in \mathcal{X} \times \mathcal{Y}$ has an underlying $p(\boldsymbol{x}|y)$ and $p(\boldsymbol{x}) > 0$. If we can estimate $p(y|\boldsymbol{x})$, any $x \in \mathcal{X}$ can be mapped to a $\hat{y}$ as: $\hat{y} = \arg\max_{y \in \mathcal{Y}} p(y|\boldsymbol{x})$. The described SMIR approach approximates the class-posterior probability $p(y|\boldsymbol{x})$.

Assuming a uniform class-prior probability $p(y) = 1/c$ the SMI has the form:

$$SMI = \frac{c}{2}\int_{\mathcal{X}} \sum_{y \in \mathcal{Y}} (p(y|\boldsymbol{x}))^2 p(\boldsymbol{x})d\boldsymbol{x} - \frac{1}{2} \qquad (9)$$

Then, by adopting an empirical kernel map:

$$\Phi_n: \mathcal{X} \to \mathbb{R}^n, x \to (k(\boldsymbol{x}, \boldsymbol{x}_1), \ldots, k(\boldsymbol{x}, \boldsymbol{x}_n))^T \qquad (10)$$

the class-posterior probability $p(y|\boldsymbol{x})$ can be approximated by:

$$q(y|\boldsymbol{x}; \boldsymbol{a}) = \langle \boldsymbol{K}^{-1/2}\boldsymbol{\Phi}_n(\boldsymbol{x}), \boldsymbol{D}^{-1/2}\boldsymbol{a}_y \rangle \qquad (11)$$

where $\boldsymbol{a} = \{\boldsymbol{a}_1, \ldots, \boldsymbol{a}_c\}$ are model parameters, $\boldsymbol{K} \in \mathbb{R}^{n \times n}$ is the kernel matrix, $\langle \cdot \rangle$ is the inner product, and $\boldsymbol{D}$ is a degree matrix. Equation (9)(9) can be written as:

$$\widehat{SMI} = \frac{c}{2n}tr\left(\boldsymbol{A}^T\boldsymbol{D}^{-\frac{1}{2}}\boldsymbol{K}\boldsymbol{D}^{-\frac{1}{2}}\boldsymbol{A}\right) - \frac{1}{2} \qquad (12)$$

where $\boldsymbol{A} \in \mathbb{R}^{n \times c}$ is the matrix representation of model's parameters. Eq. (10) regularizes a loss function $\Delta(p, q)$ that is convex w.r.t. $q$. There are three objectives: (i) minimize $\Delta(p, q)$, (ii) maximize $\widehat{SMI}$ and (iii) regularize $\boldsymbol{\alpha}$. The SMIR optimization problem is formulated as:

$$\min_{\boldsymbol{a}_1,\ldots,\boldsymbol{a}_c \in \mathbb{R}^n} \Delta(p, q) - \gamma\widehat{SMI} + \lambda\sum_y \frac{1}{2}\|\boldsymbol{\alpha}_y\|_2^2 \qquad (13)$$

where $\gamma, \lambda > 0$ are regularization parameters. If the kernel function $\boldsymbol{k}$ is nonnegative and $\lambda > \frac{\gamma c}{n}$, Eq. (9) is convex.

## IV. EXPERIMENTAL SETUP

Pixel level building detection can be seen as a traditional multiclass classification approach. The problem at hand entail to the creation of a robust classifier, capable to understand the difference of an area depicting a building from anything else, e.g. roads and vegetation. This can be achieved by incorporating meaningful information as input values and create a model capable of handling complex patterns.

### A. Dataset description

Study areas namely Area 1, Area 2 and Area 3, situated in Vaihingen city in Germany, were used for training and evaluation purposes (Fig. 1). The Area 1 mainly consists of historic buildings with notably complex building structure but also has sporadically some, often high, vegetation. The Area 2 mainly has high residential buildings with horizontal multiple planes surrounded by long arrays or groups of dense high trees. The Area 3 is a purely residential area with small detached houses that consist of sloped surfaces but there also exists a relatively low vegetation.

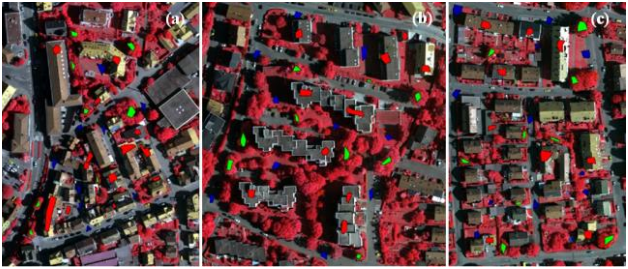

Fig. 1. Illustrating the manually annotated regions in each of the three investigated areas. Polygons in color correspond to annotations, obtained using crowdsourcing.

A Multi-Dimensional Feature Vector (MDFV) was created to feed the classifiers, as in [5]. The MDFV includes image information from the color-infrared (CIR) aerial images, the DIM point clouds and the vegetation index (Fig. 3). Additional to infrared intensities, 3D information was also considered. The cloth simulation [32] and the closest point method are applied to estimate a normalized height DIM/DSM, denoted as nDSM.

The selected parameters of the cloth simulation algorithm for all the test sites were (i) steep slope and slope processing for the scene, (ii) cloth resolution=1.5, (iii) max iterations=500 and (iv) classification threshold=0.5. This approach addresses issues related with the cost of acquisition and processing and co-registration aspects arising when data from multi-modal sources are fused together such as LIDAR/DSM and image information. Lastly, the vegetation index for every pixel $p_{ij}$ is considered and estimated through the near infrared NIR band, as:

$$NDVI = \frac{NIR - R}{NIR + R} \qquad (14)$$

where $R$ and $NIR$ refer to the red and near infrared image band. It should be mentioned that NDVI is computed only for datasets where the NIR channel is available.

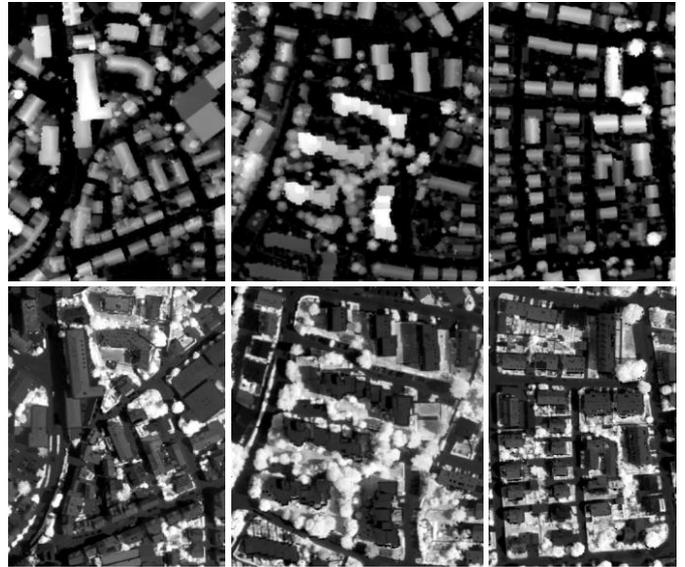

Fig. 3. Illustrating additional features for the selected areas. Top line: nDSM images, Bottom line: NDVI images.

### B. Setting up the DNN topology

A deep neural network classifier has been implemented. Fig. 2 illustrates the proposed approach. The first two encoder layer parameters, were initially set using the autoencoder approach; i.e., an unsupervised training approach, where inputs and outputs are the same [9]. Parameters for the hidden and output layers were randomly initialized. Then, a fine-tuning training step, using backpropagation algorithm, is applied to the entire network. Fine-tuning training process involved six different performance functions: standard approaches, i.e. MSE and MAE and four custom ones. The four custom performance functions are:

1. "Manifold" approach, based on the anchor graph SSL technique.

2. "SMIR" approach, based on the maximal performance Gain SSL technique.

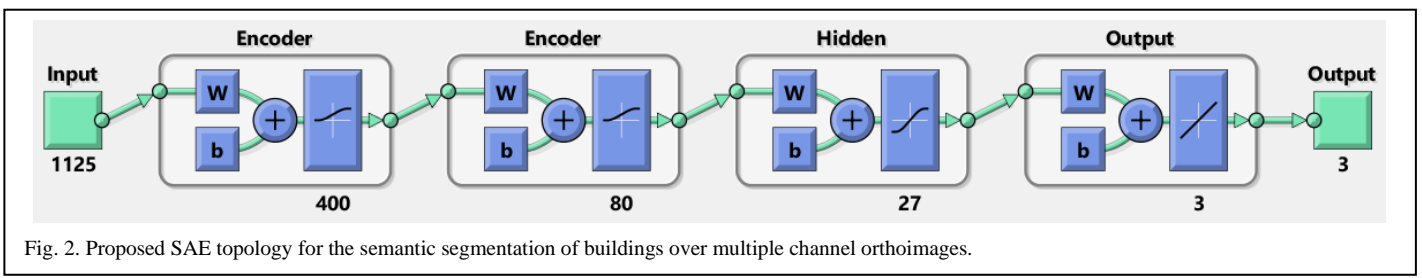

Fig. 2. Proposed SAE topology for the semantic segmentation of buildings over multiple channel orthoimages.

TABLE I. Data distribution utilized for the training process and testing process.

| | Area 1 | Area 2 | Area 3 |
|---|---|---|---|
| **Available annotated data (pixels)** | 21428 instances in 1125 dimensional space. | 19775 instances in 1125 dimensional space. | 22452 instances in 1125 dimensional space. |
| **Initial instances (pixel level) distribution** | Class  Count  Percentage<br>1  13730  64.08%<br>2  3971  18.53%<br>3  3727  17.39% | Class  Count  Percentage<br>1  9271  46.88%<br>2  4969  25.13%<br>3  5535  27.99% | Class  Count  Percentage<br>1  9205  41.00%<br>2  6003  26.74%<br>3  7244  32.26% |
| **Labeled data distribution (used for training)** | Class  Count  Percent<br>1  2197  64.05%<br>2  636  18.54%<br>3  597  17.41% | Class  Count  Percent<br>1  1484  46.87%<br>2  796  25.14%<br>3  886  27.98% | Class  Count  Percent<br>1  1473  40.98%<br>2  961  26.74%<br>3  1160  32.28% |
| **Unlabeled data distribution (used for training)** | Class  Count  Percentage<br>1  8787  64.08%<br>2  2541  18.53%<br>3  2385  17.39% | Class  Count  Percentage<br>1  5933  46.88%<br>2  3180  25.13%<br>3  3542  27.99% | Class  Count  Percentage<br>1  5891  41.00%<br>2  3842  26.74%<br>3  4636  32.26% |
| **Unseen (test) data (used for evaluation)** | Class  Count  Percentage<br>1  2746  64.08%<br>2  794  18.53%<br>3  745  17.39% | Class  Count  Percentage<br>1  1854  46.89%<br>2  993  25.11%<br>3  1107  28.00% | Class  Count  Percentage<br>1  1841  41.01%<br>2  1200  26.73%<br>3  1448  32.26% |

3. "SAFER" approach, based on the squared-loss mutual information SSL technique.
4. "Weighted average" approach, based on a harmonic mean of the previous three methods and "MSE" and "MAE" errors.

The initial image is separated into overlapping blogs of size $15 \times 15 \times 5$. The DNN classifier utilizes these 1,125 values and decides the corresponding class for the pixel at the center of the patch. The first two hidden layers are encoders, trained in an unsupervised way. They serve as non-linear mappers reducing the dimensionality of the feature space from 1,125 to 400 and then to 80. Then a hidden layer of 27 neurons perform a final mapping, allowing for the classification in one of the three pre-defined classes. Fig. 2 illustrates the proposed DNN topology.

*C. Amount of data required*

This work aims to demonstrate the usability of SSL techniques when limited data are available. Towards that direction, a crowdsourcing approach was considered. We ask the users to draw few polygons over the images. The only limitation was the number of classes. Users had to annotate, i.e. create at least one polygon, for each of the following three categories: Buildings (1), Vegetation (2) and Ground (3). The final areas are shown in Fig. 1. At this point, we need to clarify that less than 10% of the pixels are annotated. This was done on purpose, since we tried to keep the users' effort at minimum

Then, from available data within the polygons, we collect less than half of them, and use them to train/validate the classifiers. Concerning the vegetation class, trees with medium and high height are considered as "good" indicative samples. The ground class contains the bare-earth, roads and low vegetation (grass, low shrubs, etc.). The class buildings contain all the man-made building structures. To improve the classification process, shadowed areas of each class are also included. In addition, the training sample polygons are spatially created to improve representativity of each class and consider the spatial coherency of the content.

*D. Experimental results*

To purify the output of the classifier from the noisy data, initially only the building category is selected from the available classes. Thus, only the pixels associated with the buildings are extracted. The building mask is refined by post-processing. The goal of the post processing is to remove noisy regions such as isolated pixels or tiny blobs of pixels and retains local coherency of the data. Towards this, initially a majority voting technique with a radius of 21 pixels is implemented. Also, an erosion filter of a 7×7 window is applied.

The majority voting filter categorizes the potential building block with respect to the outputs of the neighboring output data. This filter addresses the spatial coherency that a building has. Since the orthoimages generated based on DSMs, the building boundaries are blurred due to mismatches during the application DIM algorithm. This affects the building results dilating their boundaries. Thus, the erosion filter was applied to "absorb" possible excessive interpolations on the boundaries of the buildings by reducing their dilated size.

Two alternative approaches were considered for the evaluation of the models' performance: i) over the polygons-bounded areas and ii) over the original annotations provided with the dataset. The former case involves the test data, as described in IV.A This is a typical multiclass classification approach. The latter case entails to a binary classification problem: buildings and non-buildings.

*1) The multiclass case*

In this scenario we evaluate the performance over the three available classes, i.e. Buildings (1), Vegetation (2) and Ground (3), given the annotated samples from the crowdsourced data. Table II demonstrates the model's performance over labeled, unlabeled and unseen data when different approaches are employed for the loss calculation, during fine tuning.

Obtained results indicate that the traditional loss functions, i.e. MSE or MAE, perform slightly worse than the proposed approaches. However, there are not clear indication on which technique is the most appropriate. An additional analysis step is, therefore, introduced and presented in next section.

TABLE II. PERFORMANCE COMPARISON AMONG FINE-TUNNING APPROACHES.

|  | Accuracy | Precision | Recall |
|---|---|---|---|
| **Labeled** | | | |
| MAE | 0.972 | 0.973 | 0.975 |
| ManifScore | 0.975 | 0.975 | 0.974 |
| MSE | 0.973 | 0.973 | 0.974 |
| SAFERScore | 0.971 | 0.974 | 0.974 |
| SMIRScore | 0.975 | 0.975 | 0.974 |
| WeiAve | 0.973 | 0.975 | 0.974 |
| **Unlabeled** | | | |
| MAE | 0.968 | 0.969 | 0.967 |
| ManifScore | 0.967 | 0.969 | 0.971 |
| MSE | 0.969 | 0.969 | 0.969 |
| SAFERScore | 0.967 | 0.970 | 0.970 |
| SMIRScore | 0.970 | 0.970 | 0.971 |
| WeiAve | 0.972 | 0.970 | 0.971 |
| **Unseen (Test)** | | | |
| MAE | 0.961 | 0.962 | 0.961 |
| ManifScore | 0.967 | 0.963 | 0.964 |
| MSE | 0.963 | 0.962 | 0.963 |
| SAFERScore | 0.965 | 0.964 | 0.964 |
| SMIRScore | 0.965 | 0.964 | 0.965 |
| WeiAve | 0.969 | 0.965 | 0.965 |

*2) The binary case*

Table III demonstrates the ranking among the proposed approaches. The highest F1 scores are achieved when the weighted average approach is adopted, for areas 1 and 2. Area 3 best score was achieved using SAFER. The same is observed for the Critical Success Index (CSI), which is another metric, more descriptive regarding a specific class description. There are also no clear outcomes regarding the second-best technique. Area 1 second best F1 score is achieved when trained using MAE (89.9%); areas 2 and 3 second best scores are achieved using SAFER (92.6%) and Manifold (Anchor Graph - 90.4%) approaches.

Fig. 4 demonstrates the DNN classifier's performance over small objects (e.g. single trees). Pixel annotation similarity exceeds 85% for all images. However, significant changes in annotations are observed over single objects typically vegetation. Generally, when the object spans less than $10 \times 10$ pixels, detection capabilities decline. This could be partially explained since most of the block pixels, i.e. $(15 \times 15) - (10 \times 10) = 125$ pixels, describe something different. The best DNN model, in this case, was trained using a weighted average as a performance function.

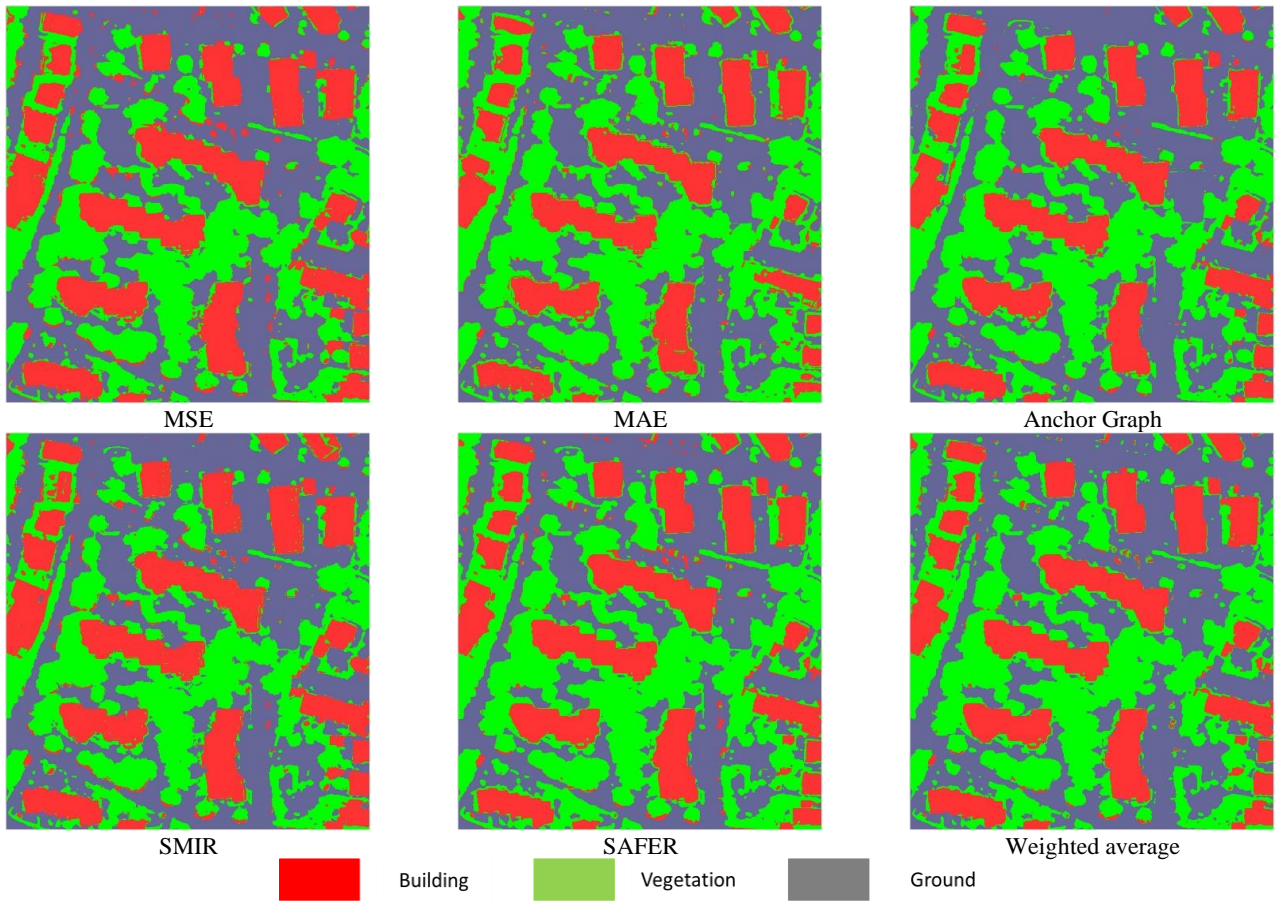

Fig. 4. Area 2 annotations for the DNN classifier, using six different performance functions for fine-tuning.

Fig. 5 illustrates the best building detection outcomes for the proposed approaches. Yellow color corresponds to pixels showing a building and model classified them as building (True Positive). Red color indicates pixels that model classified as buildings, but the actual label was either vegetation or ground (False Positive). Finally, blue color indicates areas that were buildings, but model failed to recognize them (False Negative).

TABLE III. PERFORMANCE COMPARISON FOR THE PROPOSED METHODOLOGIES IN BUILDING DETECTION. INDEX IN PARENTHESIS CORRESPONDS TO RANKING SCORE.

| Area | Performance Function | Rec (%) | Pr (%) | CSI (%) | F1 |
|---|---|---|---|---|---|
| Area 1 | MSE | 95.4 (2) | 84.9 (3) | 81.5 (3) | 89.8 (3) |
| | MAE | 94.9 (4) | 85.4 (2) | 81.6 (2) | 89.9 (2) |
| | ManifScore | 95.0 (3) | 84.3 (4) | 80.8 (4) | 89.3 (4) |
| | SMIRScore | **95.9** (1) | 83.1 (6) | 80.3 (5) | 89.0 (5) |
| | SAFERScore | 93.2 (6) | 84.3 (4) | 79.4 (6) | 88.5 (6) |
| | WeiAveTwo | 94.3 (5) | **87.1** (1) | **82.7** (1) | **90.6** (1) |
| Area 2 | MSE | **92.8** (1) | 91.8 (6) | 85.7 (3) | 92.3 (3) |
| | MAE | 85.3 (6) | **95.5** (1) | 82.0 (6) | 90.1 (6) |
| | ManifScore | 88.6 (5) | 95.1 (2) | 84.7 (5) | 91.7 (5) |
| | SMIRScore | 90.3 (4) | 94.1 (4) | 85.4 (4) | 92.2 (4) |
| | SAFERScore | 91.6 (2) | 93.7 (5) | 86.3 (2) | 92.6 (2) |
| | WeiAveTwo | 91.2 (3) | 94.6 (3) | **86.7** (1) | **92.9** (1) |
| Area 3 | MSE | 86.0 (6) | **94.9** (1) | 82.2 (4) | 90.2 (4) |
| | MAE | **91.3** (1) | 86.9 (6) | 80.3 (6) | 89.0 (6) |
| | ManifScore | 87.7 (5) | 93.3 (2) | 82.5 (2) | 90.4 (2) |
| | SMIRScore | 87.8 (4) | 92.7 (3) | 82.1 (5) | 90.2 (4) |
| | SAFERScore | 88.7 (3) | 92.6 (4) | **82.9** (1) | **90.6** (1) |
| | WeiAveTwo | 89.6 (2) | 91.1 (5) | 82.4 (3) | 90.3 (3) |

In this case, DNN classification capabilities become apparent. Segmentation for building blocks is extremely accurate considering the limited training sample. Misclassification involved inner yards, kiosk size buildings (e.g. bus stations), and the edges of the buildings. An increase of the training samples could mitigate such effects.

Table IV demonstrates the advantages of our approach against other techniques, using Critical Success Index (CSI) scores. If LiDAR is not available (i.e. the common case scenario), our approach is at par or even better compared to other techniques, despite the limited data used.

TABLE IV. COMPARATIVE RESULTS AGAINST OTHER SotA TECHNIQUES

| SotA | Data type | CSI |
|---|---|---|
| [33] | Orthoimages+LIDAR/DSM | 89.7 |
| [5] | Orthoimages+DIM/DSM | 82.7 |
| DNN (MSE) | Orthoimages+DIM/DSM | 83.2 |
| DNN (MAE) | Orthoimages+DIM/DSM | 81.3 |
| DNN (Manif) | Orthoimages+DIM/DSM | 82.7 |
| DNN (SMIR) | Orthoimages+DIM/DSM | 82.6 |
| DNN (SAFER) | Orthoimages+DIM/DSM | 82.8 |
| DNN (WeiAve) | Orthoimages+DIM/DSM | 83.9 |
| [34] | LIDAR (as point cloud) | 84.6 |
| [35] | LIDAR (as point cloud) + images | 83.5 |

V. CONCLUSIONS

Semi-supervised inspired performance functions have been utilized to fine-tune DNN for the detection of buildings over orthoimages. Results indicate that using crowdsourcing and limited train data instances, i.e. selected polygons span less than 5% of the total images area, suffice for the creation of robust detectors. The approach is based on stacked autoencoders capabilities of non-linear dimensionality reduction. That way the information of $15 \times 15 \times 5$ is summarized to 80-values feature space. The reduced dimensionality allows the implementation of SSL approaches in a way that unlabeled data can be beneficial to the model's training process. The DNN is a stacked network using the encoding layers of the stacked autoencoders. The fine-tuning performance function is designed from scratch, so that the information provided from the unlabeled data is also considered. At the end, the generated network can be easily used to any kind of new data.

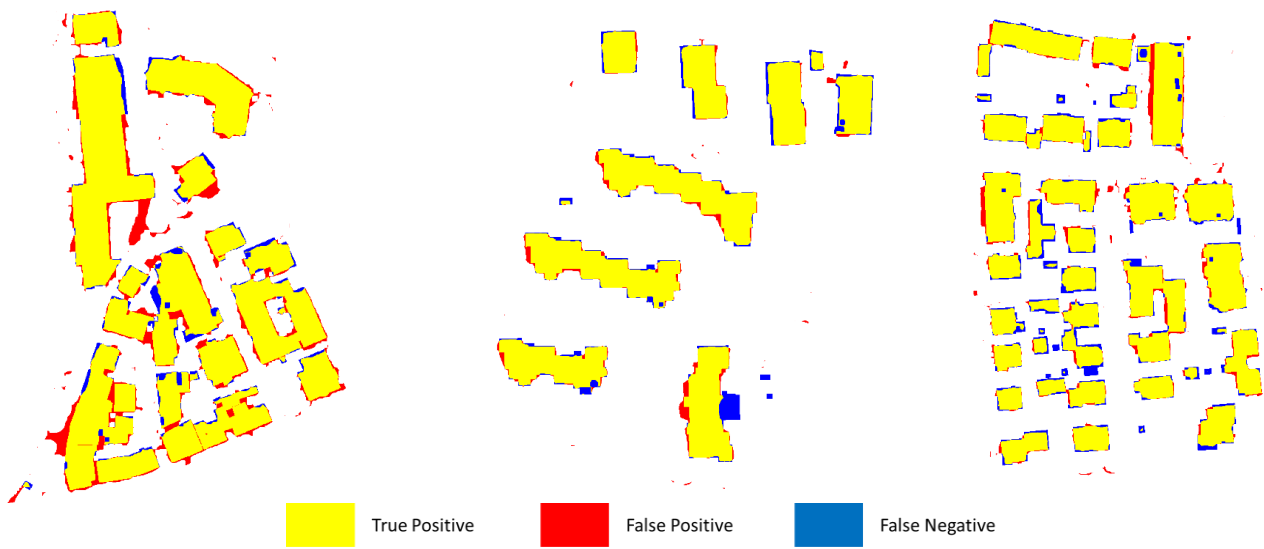

Fig. 5. Illustrating best building detection results per area. Left: Area 1 (weighted average), center: Area 2 (weighted average), right: Area 3 (SAFER).